\documentclass[sigconf,nonacm, natbib=false]{acmart}

\usepackage[ruled,vlined]{algorithm2e}
\usepackage{amsmath}

\begin{document}

\title{NEAT-NC: NEAT guided Navigation Cells for Robot Path Planning}

\author{Hibatallah MELIANI}
\orcid{0009-0006-5803-5013}
\affiliation{%
  \institution{ISI Laboratory, National School of Applied Sciences (ENSA), Abdelmalek Essaadi University}
  \city{Tetouan}
  \country{Morocco}
}
\email{meliani.hibatallah@etu.uae.ac.ma}

\author{Khadija SLIMANI}
\orcid{0000-0001-8036-2260}
\affiliation{%
  \institution{EsieaLab LDR, Higher School of Computer Science, Electronics and Automation (ESIEA)}
  \city{Paris}
  \country{France}}
\affiliation{%
  \institution{ISI Laboratory, National School of Applied Sciences (ENSA), Abdelmalek Essaadi University}
  \city{Tetouan}
  \country{Morocco}}
\author{Samira KHOULJI}
\orcid{0000-0003-2015-7913}
\affiliation{%
  \institution{ISI Laboratory, National School of Applied Sciences (ENSA), Abdelmalek Essaadi University}
  \city{Tetouan}
  \country{Morocco}
}

\renewcommand{\shortauthors}{Meliani et al.}

\begin{abstract}
To navigate a space, the brain makes an internal representation of the environment using different cells such as place cells, grid cells, head direction cells, border cells, and speed cells. All these cells, along with sensory inputs, enable an organism to explore the space around it. Inspired by these biological principles, we developed NEAT-NC, a NeuroEvolution of Augmenting Topology guided Navigation Cells. The goal of the paper is to improve NEAT algorithm performance in path planning in dynamic environments using spatial cognitive cells. This approach uses navigation cells as inputs and evolves recurrent neural networks, representing the hippocampus part of the brain. The performance of the proposed algorithm is evaluated in different static and dynamic scenarios. This study highlights NEAT’s adaptability to complex and different environments, showcasing the utility of biological theories. This suggests that our approach is well-suited for real-time dynamic path planning for robotics and games.
\end{abstract}


\maketitle

\section{Introduction}

During navigation, the brain forms an internal spatial representation that encodes the relationships among locations in the environment. This representation is called a mental map. A mental map is made with the help of place cells that fire when the organism is in a specific location and form a memory of the place \cite{o2023place} \cite{sheffield2019dendritic}. Place cells were discovered in the hippocampus \cite{o1971hippocampus} \cite{hartley2014space}. Later, other cells were discovered, such as grid \cite{hafting2008hippocampus} \cite{jacobs2013direct}, head direction \cite{taube2007head}, border \cite{boccara2010grid,lever2009boundary,solstad2008representation} and speed cells \cite{kropff2015speed}, in different parts of the brain. Place cells are activated when the organism is in a certain location of the map and fire at maximum when it is facing the goal \cite{ormond2022hippocampal}. Border cells are neurons that fire when the organism is near obstacles or edges. Head direction cells fire when the head is faced towards a certain direction. Speed cells are neurons where the firing rate depends on the running speed of the individual. All these cells, along with sensory inputs, enable an organism to navigate space around them. Based on these principles, much research has developed algorithms for autonomous robot path planning.

Autonomous robots need to collect information from the environment using sensors and process this input to successfully navigate the environment and avoid obstacles. Path planning is a core function of autonomous mobile robot technology and is one of the most critical problems in autonomous robot navigation. It requires finding a feasible obstacle-free path in an optimal amount of time, using indicators such as path length, time, and smoothness. There are mainly two types of path planning problems: static and dynamic. The first type contains only static obstacles in the environment, and the second contains both static and moving obstacles.

The path planning techniques have moved from classical deterministic methods to advanced metaheuristic, bio-inspired, and AI-based approaches due to the increased complexity of real-world problems and the need for more adaptive and flexible solutions. Some of the known algorithms are Genetic Algorithm (GA), Simulated Annealing (SA), Particle Swarm Optimization (PSO), and Ant Colony Optimization (ACO). With the rapid advancement of artificial intelligence (AI) and machine learning (ML), these techniques have been widely used to address this problem. Their adaptive nature and ability to learn from experience make them particularly well suited for dynamic and uncertain environments. Other algorithms inspired by biology were also used, such as NeuroEvolution of Augmenting Topologies (NEAT), which is an evolutionary algorithm that optimizes both the weights and structure of neural networks. Its ability to adapt network complexity over time makes it effective for reinforcement learning in dynamic and unpredictable environments.

As part of a multi-stage framework for autonomous agents, this work builds upon previous research on inverse kinematics \cite{meliani2025tempga} and addresses autonomous navigation. The paper focuses on developing a brain-inspired navigation model that draws on spatial cognition cells observed in biological neural systems.The goal of the paper is to improve the NeuroEvolution of Augmenting Topology (NEAT) algorithm by feeding inputs required to navigate a dynamic environment. Instead of forming a cognitive mental map, we use the firing of different spatial cells to help the agent navigate the environment. To our knowledge no one has integrated those biological principles into the NEAT algorithm.

In this paper, we propose an improved version of NEAT that can solve the dynamic navigation problem by adding an architecture that mimics the hippocampus and navigation cells.

In summary, the contributions of this paper are:

\begin{enumerate}

  \item A brain-inspired navigation cells representation combining goal-oriented place cells, head-direction signals, border cells and speed cell as inputs to a neuroevolutionary network.

  \item An evolutionary navigation system capable of handling dynamic obstacles relying on recurrent memory simulating the hippocampus memory structures for spatial information.

  \item A hippocampus-inspired fitness function that promotes efficient spatial navigation by rewarding straighter, goal-directed paths.

\end{enumerate}

The structure of the paper is organized as follows: First, Section ~\ref{sec:RW} presents some related works. Next, Section ~\ref{sec:Method} explains the proposed algorithm. Following that, Section ~\ref{sec:Exp} and ~\ref{sec:Result} focus on presenting the experiment and evaluating the effectiveness of the method. Finally, Section ~\ref{sec:conc} summarizes the research, draws conclusions, and discusses future perspectives. 

The code of this project is available at \url{https://github.com/HHNM/NEAT-NC}.
\section{Related Work}\label{sec:RW}

To solve the path planning problem, a wide array of methodologies has been explored, ranging from classical sampling-based algorithms to advanced hybrid and reinforcement learning frameworks. Foundational approaches often rely on geometric sampling or evolutionary heuristics to navigate complex spaces. For instance, \cite{hu2025path} improves upon classical structures by proposing a parallel sampling and bidirectional guidance Rapidly-Exploring Random Tree (PB-RRT), specifically optimized for dynamic environments. Building on evolutionary concepts, \cite{teng2025path} integrates an improved genetic algorithm with a dual-layer fuzzy control system to enhance navigation in intricate layouts.

A significant trend in recent literature is the fusion of global optimization with local obstacle avoidance to ensure both efficiency and safety. \cite{wang2025hybrid} exemplifies this by combining Modified Golden Jackal Optimization (MGJO) for global search with an Improved Dynamic Window Approach (IDWA) for local maneuvering. Similarly, \cite{liu2025fusion} utilizes a fusion of improved Gray Wolf Optimization (GSGWO) and IDWA, demonstrating the effectiveness of metaheuristic-local hybrids. Adding a layer of adaptive logic, \cite{meliani2024robot} employs adaptive Simulated Annealing refined by Fuzzy Tsukamoto, while \cite{slimani2025real} focuses on real-time responsiveness through a dynamic adaptive routing (DAR) approach.

The integration of Machine Learning has further shifted the focus toward autonomous decision-making and predictive modeling. To address the slow convergence of traditional models, \cite{maoudj2020optimal} proposes an Efficient Q-Learning (EQL) algorithm. This is further specialized by \cite{zhong2025cross}, who incorporate simulated annealing principles and heuristic rewards into a Q-learning framework to balance exploration and exploitation. Furthermore, \cite{deshpande2024mobile} proposes a hybrid reinforcement learning approach combining Deep Deterministic Policy Gradient (DDPG) with Differential Gaming (DG). Finally, \cite{stapleton2022neuroevolutionary} pushes the boundaries of trajectory prediction by using multi-objective neuroevolution (NSGA-II) to optimize hyperparameters for combined CNN and LSTM networks.

Many researchers have used NEAT to solve the path planning problem. This approach \cite{shrestha2025near} integrates NEAT with reinforcement learning and evolutionary strategies to improve policy learning and efficiency. This study \cite{zhang2025neat} proposes a path-planning framework that combines neural evolution with graph-based modeling to jointly optimize coverage completeness and path smoothness for 3D inspection tasks. In \cite{sinha2025towards} NEAT is used, with an improved reward function, to control a planar snake robot in obstacle-dense environments. This paper \cite{shrestha2025reinforced} explores NEAT for environment management, demonstrating its application in multi-room navigation using simulations of real-world scenarios.

Inspired by the brain and the hippocampus region, much research has integrated navigation cells principles in algorithms to solve path planning. Foundational models in this domain prioritize the interaction between distinct spatial cell types to anchor agents within their environments. For instance, \cite{gay2021towards} utilizes place cells to store local environment representations while employing grid and head-direction cells to predict agent positions. Building on this hierarchical structure, \cite{hu2019spatial} introduces a grid cell-based state input for reinforcement learning, constructing a multi-scale model inspired by the varying resolutions of hippocampal place cell scales. To further refine these representations, \cite{hicks2025bio} proposes a Goal-Directed Cognitive Map (GDCM) model that integrates head-direction, speed, border vector, grid, and place cells, allowing for the construction of dynamic spatial maps without requiring exhaustive exploration.

Other researchers have focused on the circuit-level logic and functional extensions of these biological units. \cite{zhang2025brain} proposes a brain-inspired path-planning algorithm that utilizes spiking neural networks (SNNs) to specifically model place cells and navigation behaviors. In a more streamlined approach, \cite{zhang2024endotaxis} develops an endotaxis neural algorithm using a simple three-layer biologically inspired circuit—comprising resource, point, map, and goal cells—to enable learning and problem-solving in complex layouts. Finally, the work in \cite{cuperlier2007neurobiologically} presents a hippocampal–prefrontal-inspired navigation model for mobile robots, introducing "transition cells" as a functional extension of standard place cells to better handle navigation tasks.

\section{Methodology}\label{sec:Method}
The proposed NeuroEvolution of Augmenting Topology guided Navigation Cells (NEAT-NC) uses navigation cells as input and feeds them to the recurrent neural network (RNN) to solve static and dynamic path planning. The algorithm uses those cells to detect goal, obstacles around the agent and decide its direction and speed. In addition, RNN acts as a spatial memory that remembers obstacles and avoids them. A fitness function is designed to encourage agents to follow confident paths while minimizing traversal time. The algorithm returns as outputs the agent’s angular and linear velocity.
\begin{figure}[h]
  \centering
  \includegraphics[width=\linewidth]{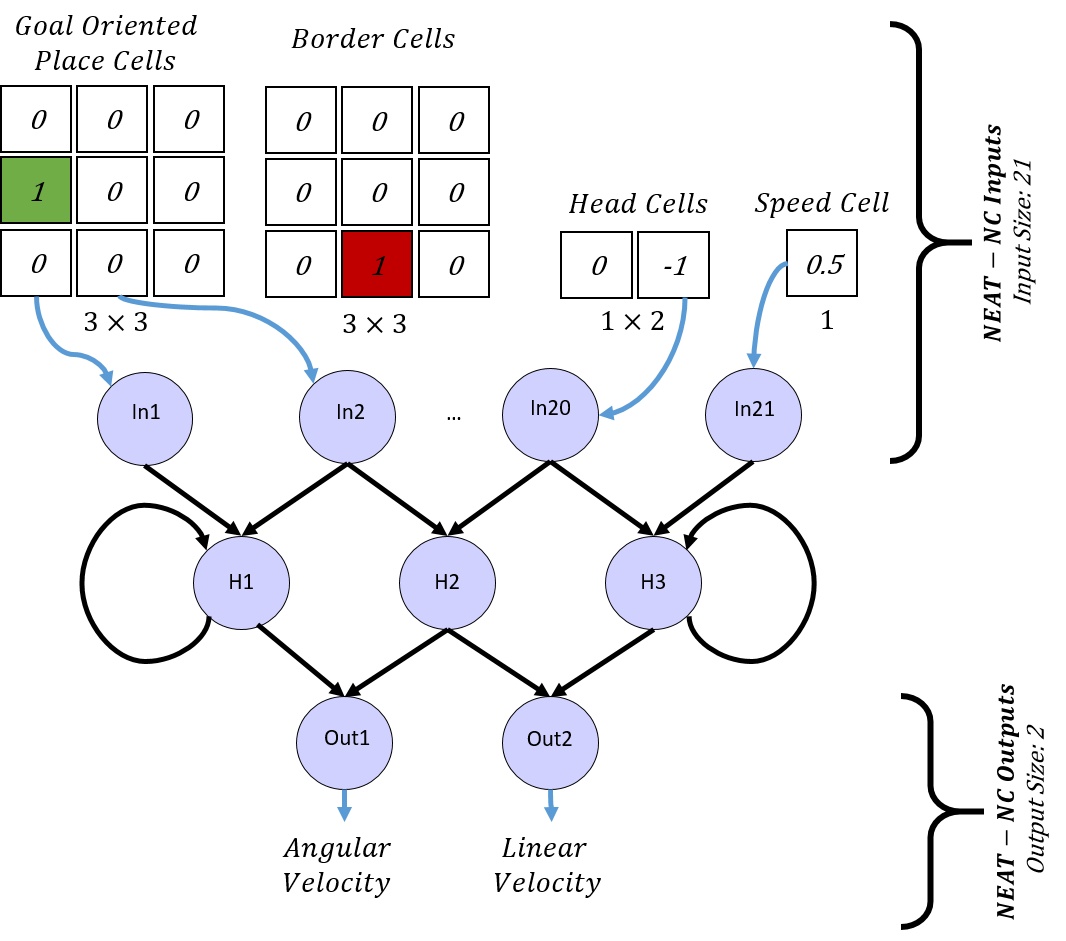}
  \caption{Topology of NeuroEvolution of Augmenting Topology guided Navigation Cells (NEAT-NC)}
  \label{fig:NEAT-NC}
\end{figure}

The topology of NEAT-NC is shown in Figure \ref{fig:NEAT-NC}, while the details of the method are described in the following subsections.

\subsection{Encoding for NEAT-NC}

\begin{figure}[h]
  \centering
  \includegraphics[width=\linewidth]{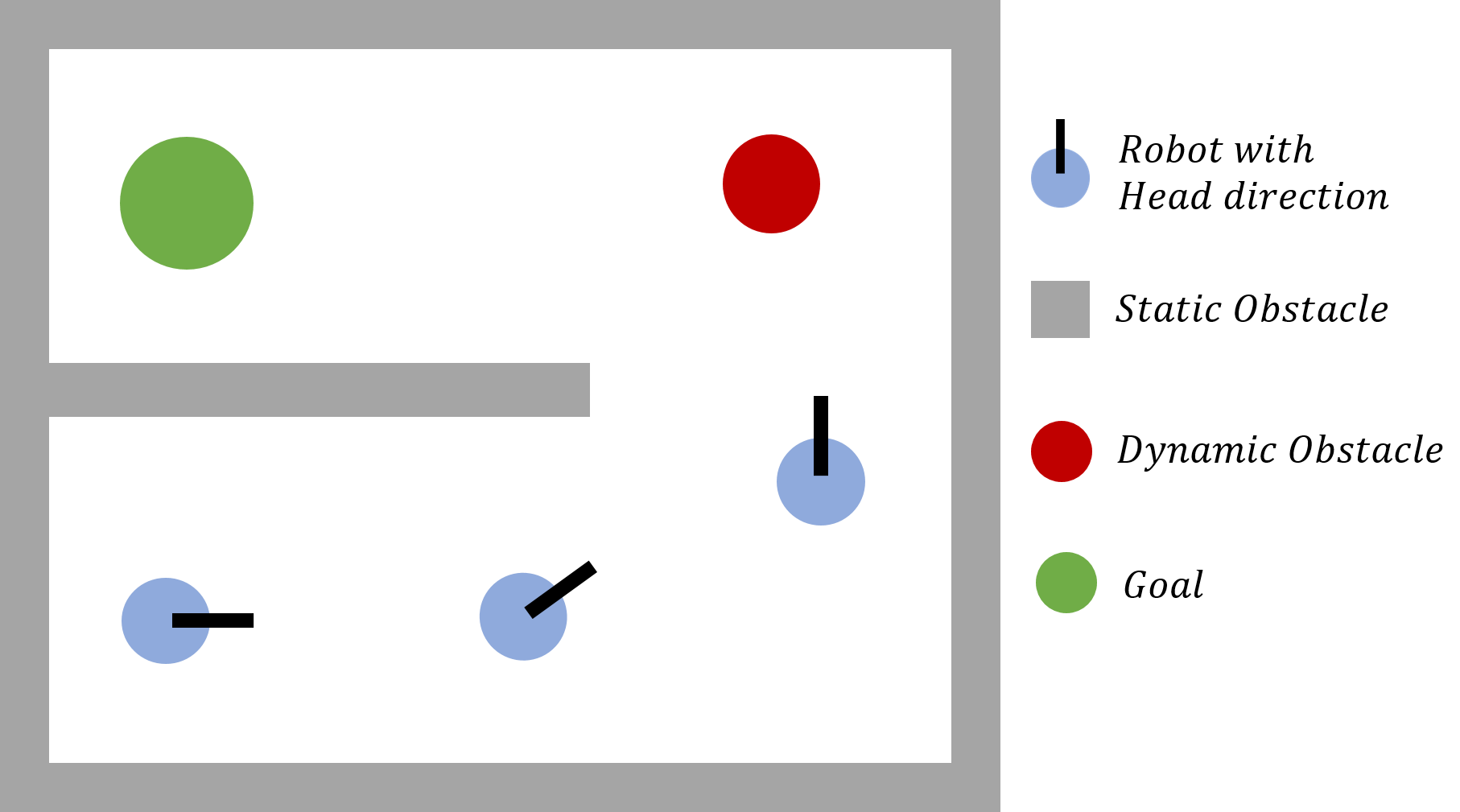}
  \caption{The Elements recognized by the navigation cells of NEAT-NC in the environment.}
  \label{fig:Env}
  \end{figure}

\begin{figure}[h]
  \centering
  
  \includegraphics[scale= 0.3]{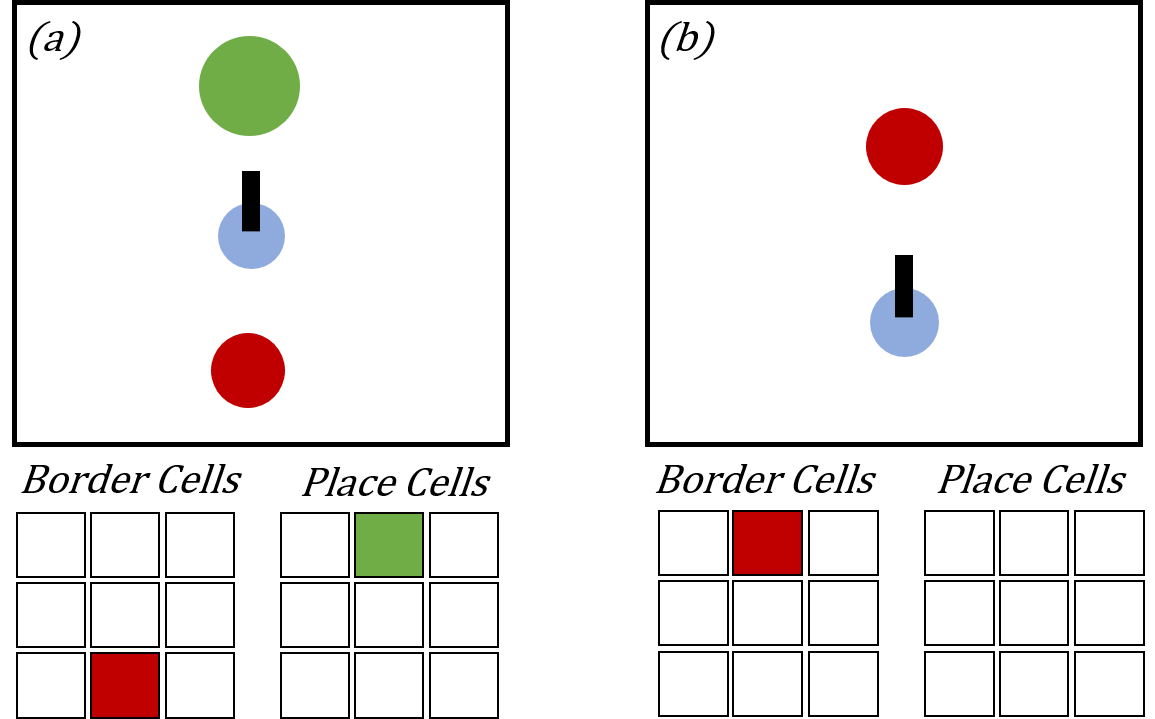}
  \includegraphics[width=\linewidth]{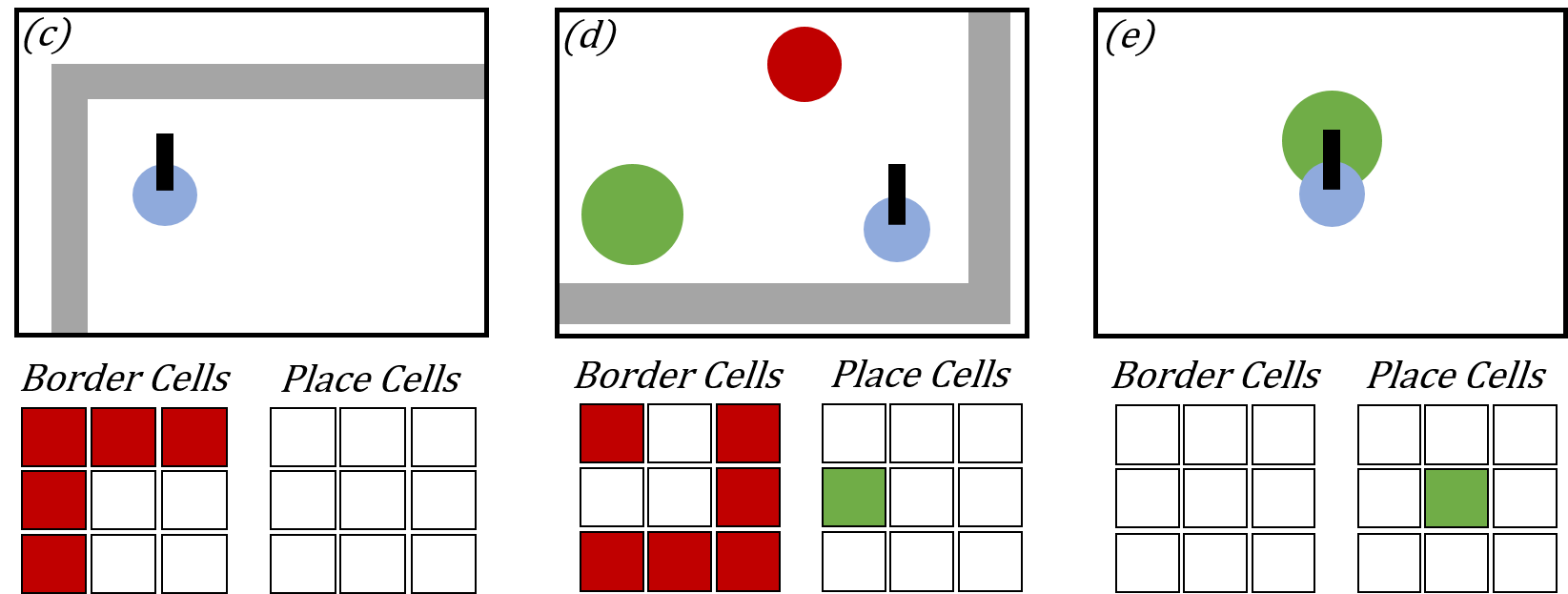}
  \caption{Place and border cells reaction in different Scenarios.}
  \label{fig:Navigation Cells}
\end{figure}

This part focuses on integrating navigation biological principles into the algorithm design. Every individual in NEAT-NC represents a Recurrent Neural Network (RNN), where the inputs are inspired by four navigation cells: goal-oriented place cells, border cells, head-direction cells, and speed cell (Figure \ref{fig:Env} and \ref{fig:Navigation Cells}), while the outputs are angular and linear velocity.


\begin{figure}[h]
  \centering
   \includegraphics[scale= 0.296]{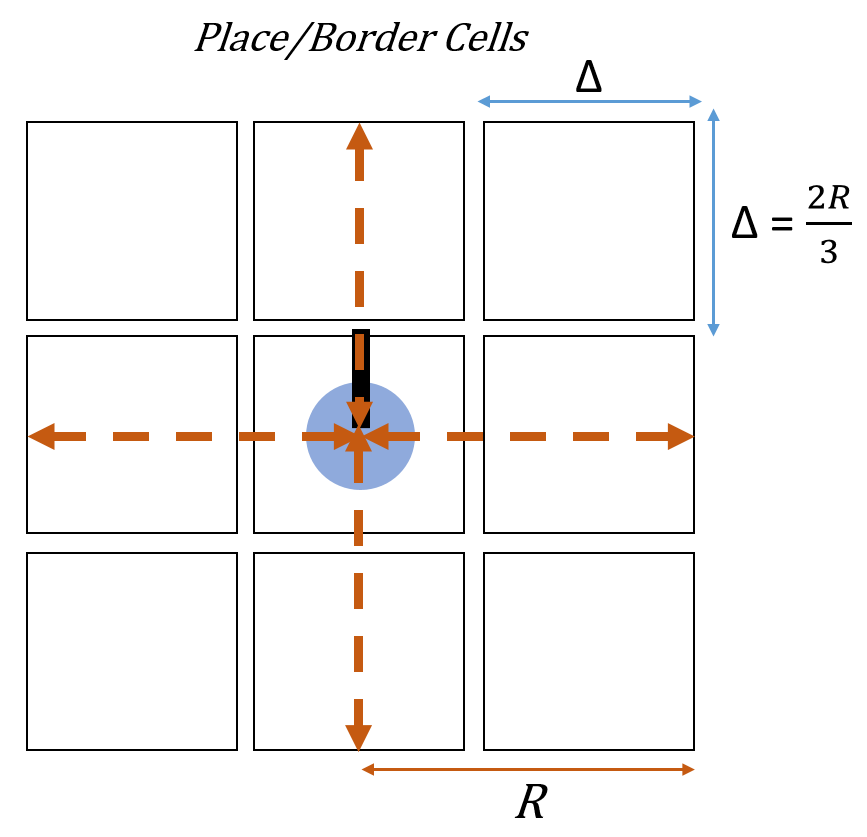}
  \includegraphics[scale= 0.296]{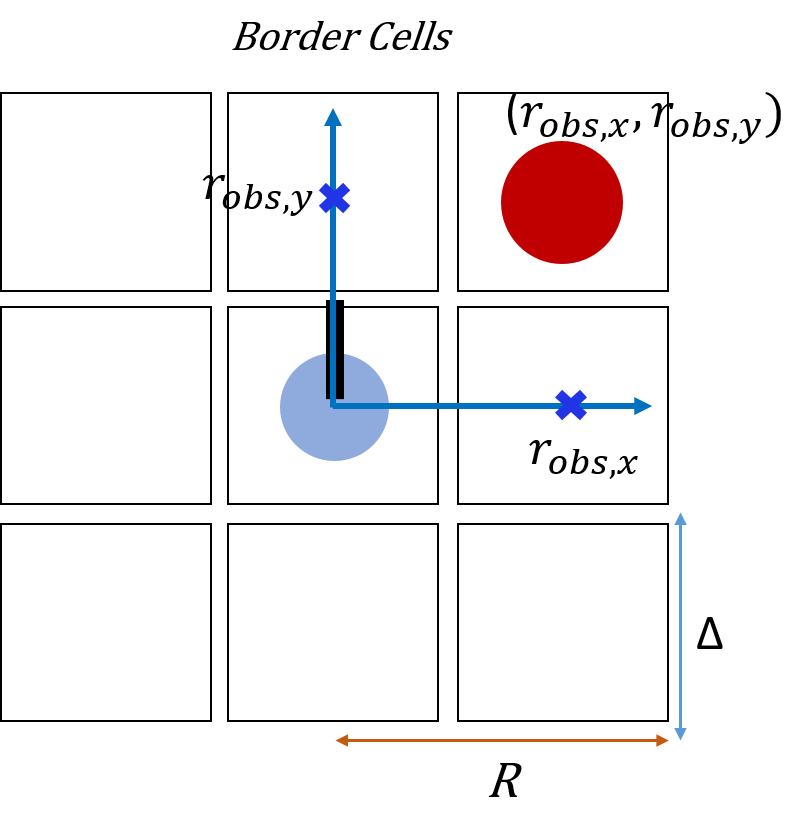}

  \caption{Border and Place cells grid placement.}
  \label{fig:BorderPlace}
\end{figure}

The input encoding of NEAT-NC consists of four layers (Algorithm \ref{alg:Cells_input}):

\noindent $\bullet$ \textbf{Border Cells:} These cells are represented by a $3\times 3$ grid surrounding the agent, where each “border cell” fires with a value of 1.0 if a static wall or dynamic obstacle at world coordinates $(x_{\text{obs}}, y_{\text{obs}})$ occupies its area within the sensor radius $R$, otherwise 0 (Figure \ref{fig:Navigation Cells} and \ref{fig:BorderPlace}). Mathematically, The value of a single cell is:

\begin{equation}
\text{border}[g_y, g_x] = 
\begin{cases} 
1, & \text{if } \sqrt{(x_{\text{obs}} - x_{\text{agent}})^2 + (y_{\text{obs}} - y_{\text{agent}})^2} \le R \\
0, & \text{otherwise}
\end{cases}
\end{equation}

\begin{equation}
g_x = \left\lfloor \frac{r_{obs,x} + R}{\Delta} \right\rfloor, \quad
g_y = \left\lfloor \frac{r_{obs,y} + R}{\Delta} \right\rfloor
\end{equation}
where $\Delta = \frac{2R}{N}$ is the cell size, $R$ is the perception radius, and $N=3$ is the grid resolution. The term $(r_{obs,x}, r_{obs,y})$ represents the obstacle coordinates transformed into the agent's frame via a the rotation matrix $R(-\theta)$ to align with the agent's heading:

\begin{equation}
\begin{bmatrix} r_{obs,x} \\ r_{obs,y} \end{bmatrix} = 
\begin{bmatrix} \cos(-\theta) & -\sin(-\theta) \\ \sin(-\theta) & \cos(-\theta) \end{bmatrix}
\begin{bmatrix} x_{\text{obs}} - x_{\text{agent}} \\ y_{\text{obs}} - y_{\text{agent}} \end{bmatrix}
\end{equation}
A grid cell $\text{grid}[g_y, g_x]$ is set to $1.0$ if the calculated indices fall within the valid bounds of the grid ($0 \le g_x, g_y < N$), indicating the presence of an obstacle within that specific spatial bin. Otherwise, the cell remains $0.0$.

\noindent $\bullet$ \textbf{Place Cells:} Represented by another $3\times 3$ grid encoding the goal’s position $(x_{\text{goal}}, y_{\text{goal}})$ relative to the agent (Figure \ref{fig:Navigation Cells} and \ref{fig:BorderPlace}). The value of the cell is set similarly to the border cells:

\begin{equation}
\text{place}[g_y, g_x] = 
\begin{cases} 
1, & \text{if } \sqrt{(x_{\text{goal}} - x_{\text{agent}})^2 + (y_{\text{goal}} - y_{\text{agent}})^2} \le R \\
0, & \text{otherwise}
\end{cases}
\end{equation}

\begin{equation}
g_x = \left\lfloor \frac{r_{goal,x} + R}{\Delta} \right\rfloor, \quad
g_y = \left\lfloor \frac{r_{goal,y} + R}{\Delta} \right\rfloor
\end{equation}

where $(r_{goal,x}, r_{goal,y})$ are the goal coordinates transformed by the rotation matrix $R(-\theta)$ to align with the agent's heading.

 \noindent $\bullet$ \textbf{Head-Direction Cells:} Encode the agent’s heading relative to the goal. Let $\theta_{\text{agent}}$ be the agent’s orientation and $\theta_{\text{goal}} = \arctan2(y_{\text{goal}} - y_{\text{agent}}, x_{\text{goal}} - x_{\text{agent}})$ the angle to the goal. Then, the relative angle is $\Delta \theta = \theta_{\text{goal}} - \theta_{\text{agent}}$, and the head-direction cells are:

\begin{equation}
\text{head\_dir} = 
\begin{bmatrix}
\sin(\Delta \theta) \\
\cos(\Delta \theta)
\end{bmatrix}, \quad 
\text{head\_dir} \in [-1, 1]
\end{equation}

\noindent $\bullet$ \textbf{Speed Cell:} Represents the agent’s normalized linear velocity. If $v_{\text{agent}}$ is the current speed and $v_{\max} = 3.0$ the maximum speed, the speed cell value is:

\begin{equation}
\text{speed\_cell} = \frac{v_{\text{agent}}}{v_{\max}} \in [0, 1]
\end{equation}

By utilizing an RNN, the architecture maintains internal hidden states that encode the temporal history of cell-based activations (Algorithm ). This recurrence is essential for navigating dynamic environments, as it enables the agent to integrate past observations, internalize the motion patterns of dynamic obstacles and maintain goal-directed behavior. This recurrent neural network generates two outputs: the first determines the agent’s angular velocity, while the second regulates linear velocity.

\begin{algorithm}
\caption{NEAT-NC Encoding}\label{alg:Cells_input}
\KwData{Agent state ($x, y, \theta, v$), Goal position $P_g$, Obstacles List $O$, Sensor radius $R$, Agent max speed $v_{max}$} 
\KwResult{Input vector $\mathcal{I} \in \mathbb{R}^{21}$}
\BlankLine
Initialize $G_{obs} \leftarrow \text{zeroMatrix}(3, 3)$\;
Initialize $G_{goal} \leftarrow \text{zeroMatrix}(3, 3)$\;
$cell\_size \leftarrow (2 \times R) / 3$\;
\BlankLine
\ForEach{Target point $P \in \{P_g\} \cup O$}{
    \If{$\text{dist}(Agent, P) \le R$}{
        $\Delta x \leftarrow P_x - x$\;
        $\Delta y \leftarrow P_y - y$\;
        $r_x \leftarrow \Delta x \cos(-\theta) - \Delta y \sin(-\theta)$\;
        $r_y \leftarrow \Delta x \sin(-\theta) + \Delta y \cos(-\theta)$\;
        $col \leftarrow \lfloor (r_x + R) / cell\_size \rfloor$\;
        $row \leftarrow \lfloor (r_y + R) / cell\_size \rfloor$\;
        \eIf{$P$ is $P_g$}{
            $G_{goal}[row, col] \leftarrow 1.0$\;
        }{
            $G_{obs}[row, col] \leftarrow 1.0$\;
        }
    }
}
\BlankLine
$\alpha \leftarrow \text{atan2}(P_{g,y} - y, P_{g,x} - x) - \theta$\;
$HeadDir \leftarrow [\sin(\alpha), \cos(\alpha)]$\;
$Speed \leftarrow v / v_{max}$\;
$\mathcal{I} \leftarrow \text{Concatenate}(G_{obs}.flat, G_{goal}.flat, HeadDir, Speed)$\;
\Return $\mathcal{I}$\;
\end{algorithm}

\subsection{Genetic Operators}

NEAT evolves networks through selection, crossover, and mutation. Based on individual fitness values, the algorithm selects individuals from the population, and elitism is applied to preserve the best solutions. The selected individuals then undergo crossover to generate offspring. The proposed algorithm uses structural mutations, specifically adding nodes and connections and weight mutations, which adjust the network's connection weights, enabling the evolution of new, more effective solutions.

\subsection{Fitness function}
The fitness function in NEAT-NC is designed to encourage efficient, biologically inspired navigation in dynamic environments. It primarily rewards progress toward the goal while penalizing unsafe or inefficient behaviors. Successful goal attainment yields a substantial terminal bonus, with additional rewards for reaching the goal in fewer steps to promote time-efficient navigation. To emulate hippocampus-inspired spatial behavior, the fitness function also incorporates a straight-path bias by penalizing excessive steering. 

For a genome \begin{math} i \end{math}, the total fitness function accumulated over an episode of length T is:
\begin{equation}
    F_i = \sum_{t=1}^{T} (r_{goal}(t) + r_{disp}(t) + r_{smooth}(t) + r_{collision}(t) + r_{see}(t))
\end{equation} 

where \begin{math} r_{goal} \end{math}, \begin{math} r_{disp} \end{math} \begin{math} r_{smooth} \end{math}, \begin{math} r_{collision} \end{math} and \begin{math} r_{see} \end{math} are the goal achievement, displacement, smoothness, collision and see goal reward respectively. 

Let the agent’s state at time step \begin{math} t\end{math}, be: 
\[
a_t = (x_t, y_t, \theta_t, v_t),
\]

where \begin{math} (x_t, y_t)\end{math} is position, \begin{math} \theta_t
\end{math} is heading angle and \begin{math} v_t \end{math} is linear velocity.

\noindent $\bullet$ \textbf{Smoothness Reward:} 
To encourage smooth, hippocampus-inspired trajectories, a penalty is applied to angular velocity.

Let \begin{math} \omega_t\end{math} be the angular velocity output of the RNN:
\begin{equation}
    r_{smooth}(t) = \lambda_\omega |\omega_t|,
\end{equation} 

where \begin{math} \lambda_\omega \end{math} controls the strength of the straightness bias and is set to -0.05.

\noindent $\bullet$ \textbf{Collision Reward:}
If the agent collides with a static or dynamic obstacle at time \begin{math} t_c \end{math}:
\begin{equation}
    r_{collision}(t_c) = \lambda_c ,
\end{equation} 
    where \begin{math} \lambda_c \end{math} = -100
    
After the collusion the episode terminates for that specific agent.

\noindent $\bullet$ \textbf{Goal Achievement Reward:} 
If the agent reaches the goal at time \begin{math} t_g \end{math}:
\begin{equation}
    r_{goal}(t_g) = \lambda_g + \lambda_s (T_{max} - t_g) ,
\end{equation} 
where \begin{math} \lambda_g \end{math} is the base success reward, \begin{math} \lambda_s \end{math}  rewards faster arrival and \begin{math} T_{max} \end{math} is the episode time limit.
\begin{math} \lambda_g \end{math}, \begin{math} \lambda_s \end{math}  and \begin{math} T_{max} \end{math} are set to 5000, 5 and 1000, respectively.

\noindent $\bullet$ \textbf{Displacement Reward:}
We added a reward based on the dot product of the agent's movement vector and the normalized vector toward the goal. This ensures that reward is granted only for effective progress along the optimal heading.

\begin{equation}
    r_{disp}(t) = (P_t - P_{t-1})\cdot{\frac{(G - P_{t-1})}{\Vert(G-P_{t-1})\Vert}} ,
\end{equation} 
Where \begin{math} P_{t} \end{math} and \begin{math} P_{t-1} \end{math} are the agent's actual and  previous position, respectively and  \begin{math} G \end{math} is the goal position. 

\noindent $\bullet$ \textbf{See Goal Reward:} 
We also introduced an additional reward when the agent’s coordinates fall within a designated area, either after the final obstacle or in the last corridor, depending on the environment, just before reaching the goal. This reward helps guide the agent to navigate the maze correctly until it reaches the point where it can "see" the goal.

\begin{equation}
    r_{see}(t) = \lambda_s ,
\end{equation} 
where \begin{math} \lambda_s \end{math} is the see zone reward and is set to 10.

\section{Experiments} \label{sec:Exp}
\begin{figure}[!htbp]
  \centering
  \includegraphics[width=\linewidth]{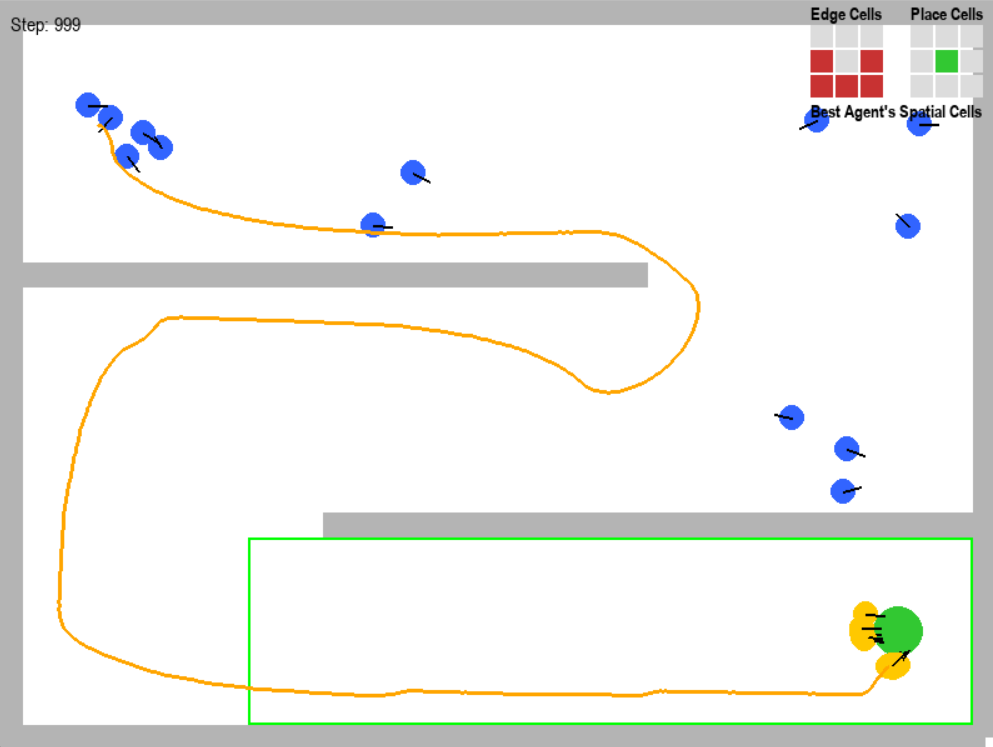}
  \caption{Environment 1 is a S maze with no dynamic obstacles}
  \label{fig:Environment1}
\end{figure}

\begin{figure}[!htbp]
  \centering
  \includegraphics[width=\linewidth]{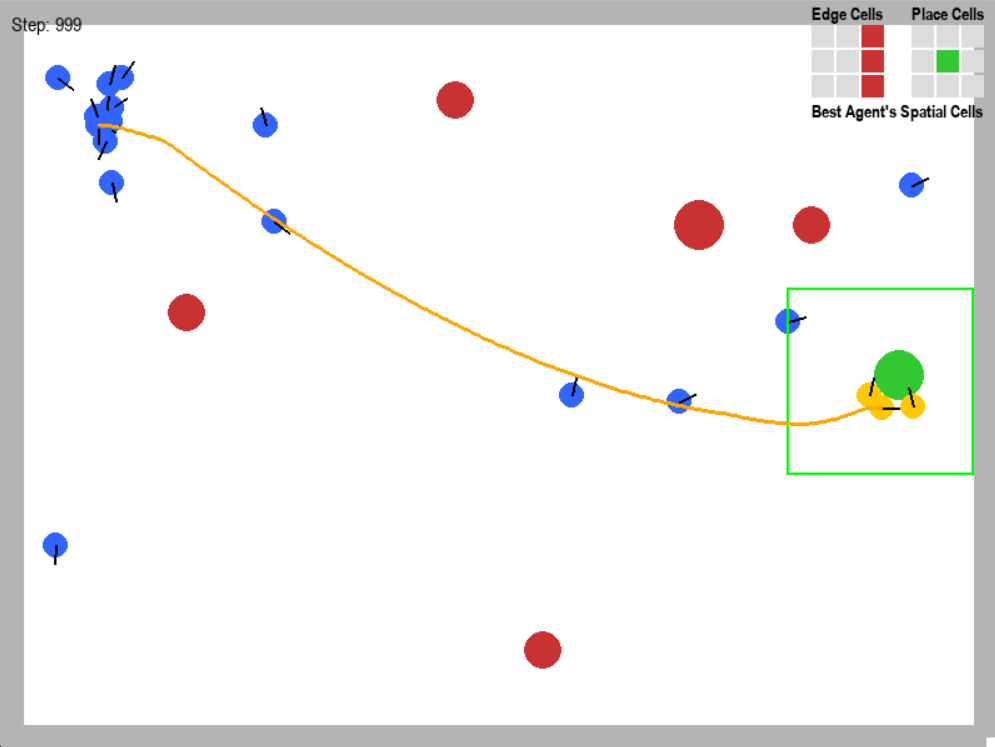}
  \caption{Environment 2 contains five dynamic obstacles}
  \label{fig:Environment2}
\end{figure}

\begin{figure}[!htbp]
  \centering
  \includegraphics[width=\linewidth]{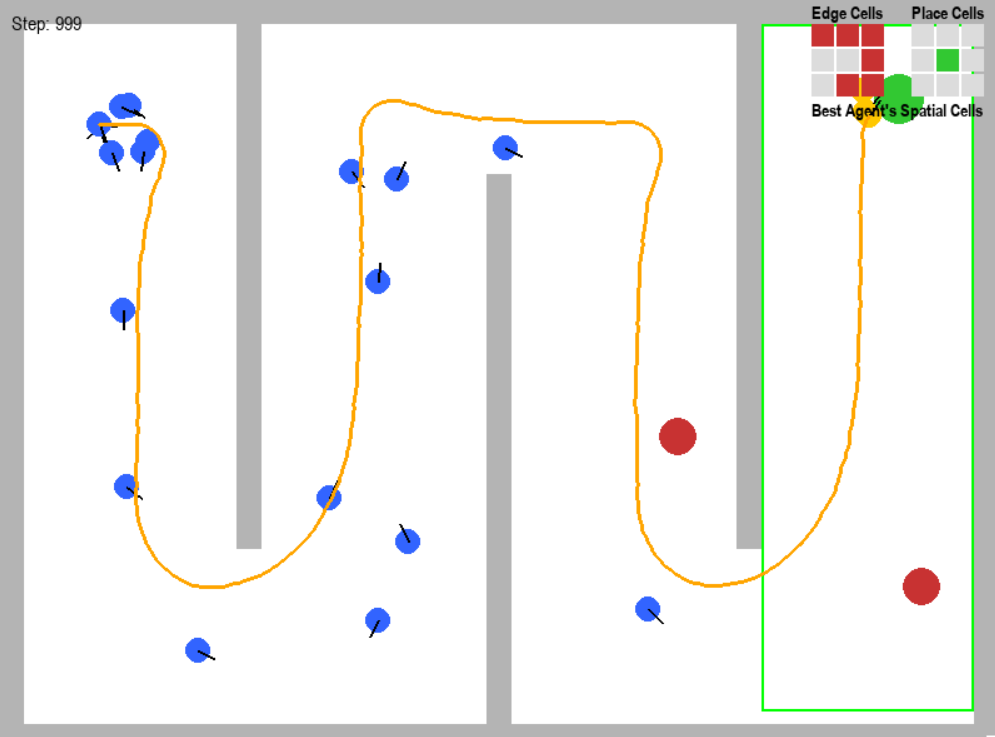}
  \caption{Environment 3 contains two dynamic obstacles}
  \label{fig:Environment3}
\end{figure}

For performance evaluation, the proposed algorithm is tested in three different scenarios with static and dynamic obstacles represented in Figures \ref{fig:Environment1}, \ref{fig:Environment2} and \ref{fig:Environment3}, varying from simple to complex environments. Dynamic obstacles, represented as red circles, move either horizontally or vertically within predefined ranges at a constant predefined velocity, introducing dynamic elements into the environment.
To ensure a fair comparison, our method is evaluated against Vanilla NEAT and Proximal Policy Optimization (PPO) was utilized as the Deep Reinforcement Learning (DRL) baseline.  The PPO agent was configured to optimize an undiscounted episodic return ($\gamma=1.0$), matching the fitness evaluation criteria of the NEAT populations. The agent utilized an Actor-Critic architecture with two hidden layers of 128 neurons each. Training was conducted over 500,000 timesteps across 30 independent runs to ensure statistical significance. A 'truncation' limit of 1,000 steps was enforced per episode, identical to the maximum lifespan of the NEAT agents, to prevent infinite loops and ensure both algorithms operated under the same temporal constraints (Table \ref{tab:ParamRL}).

\begin{table}[ht]
\centering
\caption{DRL Configuration.}
\renewcommand{\arraystretch}{1.5} 
\label{tab:ppo_condensed}
\begin{tabular}{ll}
\toprule
\textbf{Hyperparameter} & \textbf{Value / Justification} \\
\midrule
Algorithm & PPO \\
Training Steps & 500,000 \\
Episode Limit & 1,000 steps  \\
Discount ($\gamma$) & 1.0 \\
GAE ($\lambda$) & 1.0 \\
Network architecture & [128, 128] MLP \\
$n$\_steps / Batch & 4,096 / 128 \\
Learning Rate & $3 \times 10^{-4}$ \\
\bottomrule
\end{tabular}
\label{tab:ParamRL}
\end{table}

The Vanilla NEAT uses a a feedforward neural network. Both NEAT algorithms proceed to the next generation after 1,000 steps. Both NEAT and PPO agents receive the same observation space consisting of eight radar sensors. Each radar provides a normalized distance measurement to the nearest obstacle within a fixed sensing radius, resulting in an 8-dimensional continuous input vector.

The parameters for Vanilla NEAT were selected based on \cite{shrestha2025reinforced}. Table \ref{tab:Param} presents the parameters used for NEAT-NC and Vanilla NEAT. The performance of each solution was evaluated based on four criteria: success rate, fitness value, path length, and time of execution.

\begin{table}[ht]
\centering
\small
\renewcommand{\arraystretch}{1.5} 
\caption{Parameters of NEAT-NC and NEAT.}
\begin{tabular}{lccc}
\hline
\textbf{Parameters} & \textbf{NEAT-NC} & \textbf{NEAT} \\
\hline
Population size      & 50 & 50  \\
Generation      & 10 & 10  \\
Elitism      & 4 & 3  \\
Connection add rate      & 0.5 & 0.5  \\
Connection delete rate & 0.2 & 0.3  \\
Node add rate      & 0.2 & 0.2  \\
Node delete rate  & 0.2 & 0.1 \\
Weight mutate rate      & 0.8 & 0.8 \\
Fitness criterion      &  max &  max  \\
Activation function      &  Tanh &  Tanh  \\
Activation options      &  Tanh Relu Sigmoid &  Tanh Relu Sigmoid  \\
\hline
\end{tabular}
\label{tab:Param}
\end{table}

\section{Results and Discussions} \label{sec:Result}

The algorithms were tested on every instance 30 times using Python language, neat library for NEAT-NC and Vanilla NEAT, Stable Baselines3 for the DRL PPO implementation. The simulation environment was developed using Gymnasium and pygame. The computations were performed on a PC with an AMD Ryzen 7 4800H 2.90 GHz processor and 16.0 GB RAM.

Table \ref{tab:performance_metrics} reports the performance comparison of Vanilla NEAT, the proposed NEAT variant, and DRL for solving the path-planning problem in the three environments. The reported metrics include average fitness, path length, execution time, and success rate.

\begin{table}[!htbp]
\renewcommand{\arraystretch}{1.5} 
\centering
\caption{Performance metrics of the proposed and benchmark algorithms.}
\begin{tabular}{c c c c c c}
\hline
 Env. & Algorithms &  Fitness & Path & Time(s) & Success \\
 \hline

 \begin{tabular} [c]{@{}c@{}} 1\end{tabular}
 & 
 \begin{tabular} [c]{@{}c@{}} NEAT-NC \\ NEAT \\ DRL \end{tabular} & \begin{tabular} [c]{@{}c@{}} \textbf{9208.49} \\ 5999.04 \\ 6226.21 \end{tabular} & 
 \begin{tabular} [c]{@{}c@{}} \textbf{1900.63} \\ 2049.02\\  2048.65
\end{tabular} &
 \begin{tabular} [c]{@{}c@{}} \textbf{176.20} \\ 288.57 \\  1489.489
 \end{tabular} &
  \begin{tabular} [c]{@{}c@{}} \textbf{93.33\%} \\ 66.33\% \\  63.33\%
\end{tabular}
 \\
 \hline
 \begin{tabular} [c]{@{}c@{}} 2 \end{tabular} &
  \begin{tabular} [c]{@{}c@{}} NEAT-NC \\ NEAT \\ DRL \end{tabular} & \begin{tabular} [c]{@{}c@{}} \textbf{10577.67} \\ 4860.54 \\ 1781.44\end{tabular} & 
 \begin{tabular} [c]{@{}c@{}} 832.66 \\ \textbf{762.46}\\ 1409.18\end{tabular} &
 \begin{tabular} [c]{@{}c@{}} \textbf{128.77} \\ 218.98 \\ 1040.66\end{tabular} &
  \begin{tabular} [c]{@{}c@{}} \textbf{100\%} \\ 47\% \\ 16.67\%\end{tabular} 
 \\
 \hline
  \begin{tabular} [c]{@{}c@{}} 3 \end{tabular} &
  \begin{tabular} [c]{@{}c@{}} NEAT-NC \\ NEAT \\ DRL \end{tabular} & \begin{tabular} [c]{@{}c@{}} \textbf{6267.78} \\ 2281.24 \\ 798.54\end{tabular} & 
 \begin{tabular} [c]{@{}c@{}} \textbf{2120.53} \\ 2259.63\\ \textbf{2072.86}\end{tabular} &
 \begin{tabular} [c]{@{}c@{}} \textbf{186.89} \\ 307.27 \\ 1475.03\end{tabular} &
  \begin{tabular} [c]{@{}c@{}} \textbf{70.00\%}
 \\ 23.33\% \\ 6,67\%\end{tabular} 
 \\
 \hline
\end{tabular}
\label{tab:performance_metrics}
\end{table}

The Kruskal–Wallis test at a 0.05 significance level was used to detect overall performance differences among algorithms in fitness and path length, while Chi-square test was used for success rate. The significance of performance variations is quantified using p-values, which are summarized in Table \ref{tab:p-value}. When significant differences were observed, the Dunn test was performed as post-hoc analyses.

\begin{table}[h!]
\centering
\renewcommand{\arraystretch}{1.5} 
\caption{p-value for the path planning results.}
\begin{tabular}{c c c c }
\hline
 Env. & Fitness & Path & Success \\
 \hline

 \begin{tabular} [c]{@{}c@{}} 1\end{tabular}
 & 4.07e-11 & 1.74e-4 & 0.01\\
 \hline
 \begin{tabular} [c]{@{}c@{}} 2 \end{tabular} 
 & 1.85e-09 & 1.79e-06 & 1.60e-08\\
 \hline
  \begin{tabular} [c]{@{}c@{}} 3 \end{tabular}
 & 3.01e-08 & 0.042 & 4.79e-07\\
 \hline
\end{tabular}

\label{tab:p-value}
\end{table}

In terms of solution quality, evaluated through average fitness values, NEAT-NC consistently demonstrates superior performance compared to Vanilla NEAT and the DRL baseline in all environments. 
Regarding path efficiency, the proposed NEAT generates shorter paths, whereas Vanilla NEAT and DRL often produce longer paths.
The proposed NEAT approach achieves a consistently higher success rate compared to Vanilla NEAT and DRL, indicating improved robustness in navigating dynamic and complex maze structures.

\begin{figure}[h]
  \centering
  \includegraphics[width=\linewidth]{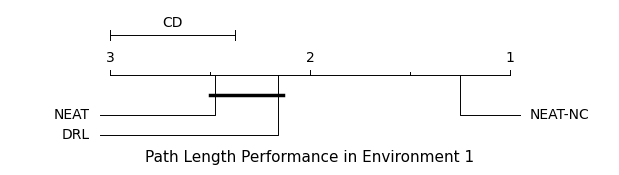}
  \includegraphics[width=\linewidth]{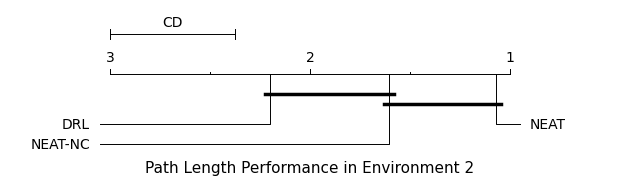}
  \includegraphics[width=\linewidth]{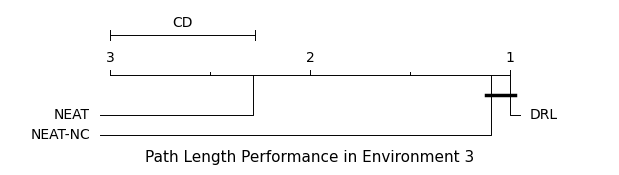}
  \caption{Dunn test's Critical Difference (CD) diagrams on Path Length}
  \label{fig:DunnP}
\end{figure}

\begin{figure}[h]
  \centering
  \includegraphics[width=\linewidth]{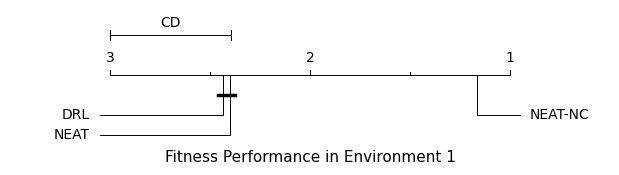}
  \includegraphics[width=\linewidth]{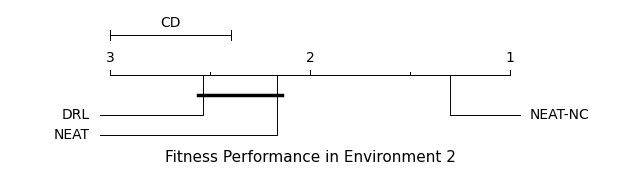}
  \includegraphics[width=\linewidth]{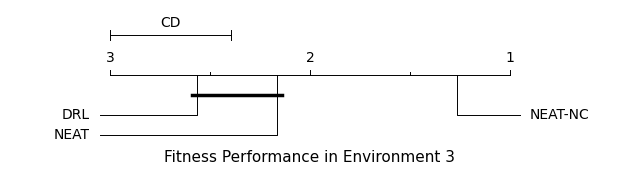}
  \caption{Dunn test's Critical Difference (CD) diagrams on fitness value}
  \label{fig:DunnF}
\end{figure}

In terms of computational performance, the proposed NEAT requires less computation time compared to vanilla NEAT. DRL generally requiring longer training times due to policy optimization and replay overhead. 

Overall, the results confirm that the proposed NEAT framework outperforms Vanilla NEAT and DRL for autonomous path planning. This highlights the effectiveness of the proposed enhancements in guiding evolutionary search toward reliable and efficient navigation behaviors.

The low p-values (<0.05) in Table \ref{tab:p-value} indicate that the performance differences between NEAT-NC and other algorithms are statistically significant. This numerical ranking was validated using the Dunn post-hoc test, confirming the statistical relationships among the algorithms. The Critical Difference (CD) diagrams (Figures \ref{fig:DunnP} and \ref{fig:DunnF}) show that NEAT-NC consistently ranks among the top-performing algorithms, indicating that the algorithm is a statistically superior algorithm to NEAT and DRL.

\section{Conclusion} \label{sec:conc}

This paper presented a brain-inspired navigation framework using the  NEAT-guided Navigation Cells (NEAT-NC) architecture to evolve a Recurrent Neural Network (RNN). The algorithm uses place, border cells, head direction cells and speed cell as input for the RNN, effectively mimicking the spatial mapping capabilities of biological systems. NEAT-NC successfully navigates different types of environments, improving success rate, path length, and speed in path planning problems in static and dynamic environments. The findings highlight the potential of integrating biological theories into algorithm design. This work complements our previous study on 7-DOF inverse kinematics by introducing a cognitive framework for autonomous navigation. Future research will focus on more advanced navigation models and the integration of navigation and manipulation into a fully interactive agent operating in a 3D environment.

\bibliographystyle{unsrt}
\bibliography{MyRef}

\end{document}